\documentclass[letterpaper, 10 pt, conference]{ieeeconf}  

\IEEEoverridecommandlockouts                              

\usepackage{xspace}
\usepackage{hyperref} 
\usepackage{mathrsfs}
\usepackage{amsmath, amssymb, amsfonts}
\usepackage[flushleft]{threeparttable}
\usepackage{tablefootnote}
\usepackage{multirow}
\usepackage{hhline}
\usepackage{graphicx}
\usepackage{subfigure}
\usepackage{hanging}
\usepackage{color}
\usepackage{tikz}
\usepackage{wrapfig}
\usetikzlibrary{calc,arrows,decorations.markings}
\usepackage{balance}
\usepackage{makecell, booktabs}
\usepackage{hhline}
\usepackage{ifthen}
\usepackage{caption}
\usepackage{graphicx}
\usepackage{booktabs}
\usepackage{graphics} 
\usepackage{epsfig} 
\usepackage{mathptmx} 
\usepackage{times} 

\let\labelindent\relax			
\usepackage{enumitem}

\newcommand{\netName}{KGNv2\xspace}

\newcommand{\SE}[1]{SE({#1})}
\newcommand{\SO}[1]{SO({#1})}
\newcommand{\trl}{T}
\newcommand{\rotMat}{R}

\newcommand{\grasp}{g}

\newcommand{\graspWidth}{w}

\newcommand{\pt}{p}

\newcommand{\gCenter}{c}

\newcommand{\height}{H}
\newcommand{\width}{W}

\newcommand{\ori}{m}
\newcommand{\oriNum}{M}

\newcommand{\centerHeatmap}{Y}
\newcommand{\centerKptsOffset}{O}
\newcommand{\scaleMap}{S}
\newcommand{\graspWidthMap}{W}

\newcommand{\loss}{L}

\newcommand{\Real}{\mathbb{R}}

\newcommand{\norm}[1]{\|{#1}\|}

\newcommand{\gauss}{\mathcal{N}}

\newcommand{\PnP}[1][]{\ifthenelse{\equal {#1} {}}{P\textit{n}P}{P{#1}P}\xspace}

\DeclareRobustCommand{\bluecheck}{%
  \tikz\fill[scale=0.4, color=blue]
  (0,.35) -- (.25,0) -- (1,.7) -- (.25,.15) -- cycle;%
}

\graphicspath{{./},{figs}}

\title{\LARGE \bf
\netName: Separating Scale and Pose Prediction for Keypoint-based 6-DoF Grasp Synthesis on RGB-D input
}

\author{Yiye Chen$^{1}$, Ruinian Xu$^{1}$, Yunzhi Lin$^{1}$, Hongyi Chen$^{1}$, and Patricio A. Vela$^{1}$%
\thanks{$^{1}$ Y. Chen, R. Xu, Y. Lin, H. Chen, and P.A. Vela are with the
School of Electrical and Computer Engineering, and the
Institute for Robotics and Intelligent Machines, Georgia Institute of
Technology, Atlanta, GA. 
{\tt\small \{yychen2019, rnx94, ylin466, hchen657, pvela\}@gatech.edu}}%
}

\begin{document}

\maketitle
\thispagestyle{empty}
\pagestyle{empty}

\begin{abstract}

We propose a new 6-DoF grasp pose synthesis approach from 2D/2.5D input based on keypoints.
Keypoint-based grasp detector from image input has demonstrated promising results in the previous study, 
where the additional visual information provided by color images compensates for the noisy depth perception.
However, it relies heavily on accurately predicting the location of keypoints in the image space. 
In this paper, we devise a new grasp generation network that reduces the dependency on precise keypoint estimation.
Given an RGB-D input, our network estimates both the grasp pose from keypoint detection as well as scale towards the camera.
We further re-design the keypoint output space in order to mitigate the negative impact of keypoint prediction noise to Perspective-n-Point (\PnP) algorithm.
Experiments show that the proposed method outperforms the baseline by a large margin, validating the efficacy of our approach.
Finally, despite trained on simple synthetic objects, our method demonstrate sim-to-real capacity by showing competitive results in real-world robot experiments.
Code is available at:~\url{https://github.com/ivalab/KGN}.

\end{abstract}

\section{Introduction}

Robotic grasping is a fundamental yet demanding problem, requiring both object perception as well as geometric reasoning based solely on sensor input.
Past reasearchers simplified the problem by constraining the grasp poses into \SE2 space, assuming that the camera looks at the scene vertically from the top,
and the gripper reaches perpendicularly to the support plane \cite{dexnet2_2017, deepgrasp2018, gknet2021}.
The restriction allows the planar grasp methods to represent grasps as simple oriented rectangles or keypoints in the image space, 
which permits directly adopting existing data-driven tools from computer vision tasks, such as object \cite{fasterrcnn2015} or keypoint \cite{CenterNet2019} detectors.
However, it also neglects possible grasp poses reaching from other directions, which potentially impedes its utility in constrained environments \cite{collision2021, graspInPrinter2021}.

The limitation of planar grasp assumption has motivated recent exploration of 6-DoF grasp synthesis, which allows grasp poses in full \SE3 space. 
Point-cloud-based methods, utilizing point set feature extractors like PointNets \cite{Pointnet2017, Pointnet++2017}, have achieved great success in generating or evaluating 6-DoF grasp poses directly from depth sensor data. 
However, these methods face empirical limitations such as poor grasp poses for small-scale objects due to limited point perception \cite{scaleBalanced}, and compromised performance in the presence of sensor noise. 
A point sampling strategy \cite{scaleBalanced} has been proposed to balance object scales, but it increases computational cost due to the need of an additional instance segmentation network.
And the vulnerability to input disturbance remains a concern.

Consequently, the utilization of 2D/2.5D input for 6-DoF grasp detection has gained attention due to the additional visual information offered by color images.
The visual clue provided by RGB modality, which can be effectively extracted by modern convolutional neural networks (CNN), can not only facilitate discerning small objects that are negligible from depth point cloud input, but also improves robustness against noise in depth sensor \cite{rgbGrasp2021}.
Despite demonstrating promising results, existing methods \cite{rgbGrasp2021}, \cite{monoGrasp} still utilize a 3D representation of grasp poses, 
necessitating the network to estimate 3D information from 2D input.
As a result, expensive annotation, such as surface normal, is needed for training \cite{monoGrasp}, or heavy distretization of \SO3 space is required \cite{rgbGrasp2021}.

To avoid the need for directly estimating 3D parameters, Keypoint-GraspNet (KGN) \cite{KGN} seperates the 2D-to-3D recovery stage out of the network.
Instead of using a 3D representation, KGN represents a grasp pose as a set of gripper keypoints in the image space and recovers the \SE3 pose from the 2D keypoints with the \PnP algorithm \cite{ippe}.
KGN avoids discretization error, as keypoint coordinates are continuous in the image space, and removes the requirement for estimating surface normal directions.
However, imprecise keypoint proximity prediction causes unstable estimation of the \textit{scale factor}, which is the magnitude of translation of a grasp pose from the camera, especially in a novel test domain, such as training on synthetic data and testing on real-world data. 
KGN heuristically addresses the issue by adopting the perceived depth as scale, which reduces the pose accuracy due to occlusion and depth sensor noise.

In this paper, we introduces \netName, an improved keypoint-based grasp detection network that enhances the accuracy of grasp pose detection. 
The network gets around the above issue by predicting pose and scale separately, which eliminates the need for accurate keypoint proximity estimation and improves the accuracy of generated pose. 
The keypoint output space is re-designed by normalizing with estimated scale, which further enhances the precisioin of estimated pose. 
The proposed modifications are simple yet effective, greatly improving performance on the primitive shape dataset from \cite{KGN}, and the network can generalize to realistic objects with significant shape variations in real-world experiments, indicating the potential of training grasp detectors on virtual data with primitive geometries, where obtaining ground-truth labels is easier.


\section{Related Work}\label{sec:Lit}
In this paper, we narrow our review on literature of learning-based 6-DoF grasp detection with anti-padel end-effector.
Other related areas include dexeterious grasp detection, planar grasp synthesis, and model-based grasp detection.
They have been thoroughly reviewed by other survey papers \cite{surveyLrn}, \cite{surveyDeep} and are outside the scope of this paper.

\subsection{Point-Cloud Methods}
With the emergent deep point set encoders such as PointNets \cite{Pointnet2017}, \cite{Pointnet++2017} and DeCo \cite{DeCo2021},
the majority of literature explores detecting 6-DoF grasp poses from point cloud input \cite{pointnet++grasp2020}.
Early effort employs a generate-then-evaluate process, where a discriminative model to predict grasp outcome is necessary \cite{GeoEval2018}. 
PointNetGPD \cite{pointnetGPD2019} uses a geometry-based heuristic approach \cite{GPG2018} to sample from \SE3 spaces, and trains a network to process enclosing point sets to score the grasp pose.
But the insufficient quality of sampled grasp poses limits the overall performance.
6DoF-GraspNet \cite{6dofGraspNet2019} replaces the sampling-based candidate proposal approach with deep generator trained with Generative Adversial Network (GAN) or Vairational Autoencoder (VAE) objective.
Other approaches \cite{refineScore2017}, \cite{refineProb2020} investigate refining the initial pose proposal by increasing the score estimated by a learnt evaluator.

The above pipeline is time-consuming due to multiple forward passes of point-cloud networks.
Driven by large scale grasping datasets \cite{acronym2021}, \cite{graspnet1b2020}, 
recent approaches turn their attention to end-to-end grasp detection - with both grasp pose parameter and confidence estimated by a single model.
The key difference lies in the grasp representation choice.
S4G \cite{s4g2020} chooses the \SE3 representation, and directly regresses the rotational and translational parameters anchored on point with high confidence.
To enable multiple detection per point, GDN \cite{GDN2020} extends the idea with a coarse-to-fine representation idea, where they first perform classificaiton on a set of discrete angular grids, and then regress translation and rotation refinement values for high-confidence candidates.
Another line of approach argues in favor of explicit contact physics reasoning.
They adopt the representation of two contact points plus pitching angle \cite{ContactGraspNet2021} \cite{L2G2022}, where the assumption is at least one contact point should be visible from partial point cloud.

Point-cloud methods share some common drawbacks, which are studied by recent literature.
Due to the high processing time to extract geometric information from the coordinates enumeration, truncating input point volume, by either downsampling \cite{s4g2020} or target segmentation \cite{6dofGraspNet2019}, is necessary. 
L2G \cite{L2G2022} alternatively designs an learnable sampler, which can be jointly tuned in the end-to-end training process.
It assumes properly designed sampling procedure can retain critical information for grasp synthesis, which is not always the case especially for high-resolution input.
Another limitation for point-cloud methods is their bias towards larger objects due to dominating point number.
Ma et al. \cite{scaleBalanced} allieviate the issue by balanced sampling based on instance segmentation mask, which relies on external segmentation module that increases computational cost.

\subsection{Image-based Methods}
Contrary to point cloud, images are faster to process with modern networks and are explicit with visual relationship, which can address the above issues. Hence, several recent research start to study 6-DoF grasp detection from image input \cite{CGPN}.
It is shown that the color modality can also improve the robustness against depth noise \cite{rgbGrasp2021}.
However, the intial exploration still relies on 3D representaion of grasp poses, such as contact-point-based description \cite{monoGrasp}, which fails to leverage existing knowledge about 3D-to-2D camera projection.

KGN \cite{KGN}, on the other hand, adopts 4 keypoints in the image space to describe a 3D grasp pose, in which the keypoints can be efficiently synthesized by well-studied keypoint detector such as CenterNets \cite{CenterNet2019}, \cite{duan2019centernet}.
Designing them to be the projection of virtual 3D points in the gripper frame with predefined coordinates, a \PnP algorithm is applied to recover 3D pose from 2D keypoints leveraging well-established camera projection principle.
Given fixed 3D coordinates, the \textit{relative location} and \textit{absolute distance} between 2D keypoints determine the \textit{pose up to a scale} and \textit{scale factor}, respectively (e.g. closer keypoints means the gripper frame is further to the camera plane).
However, the prediction of keypoint proximity is found to be unstable under novel test environment \cite{KGN}.
In this work, we relax the requirement of precise distance prediction by separately predicting the scale.
Based on the estimated scale, we further redesign the keypoint output space factoring in the influence of prediction noise on the relative location.
We show that our modified network, named \netName, achieves great performance gain.



\begin{figure*}[t!]
 
  \vspace*{0.025in}
  \centering
  \scalebox{0.95}{
    \begin{tikzpicture}
     \node[anchor=north west] at (0in,0in)
      {{\includegraphics[width=0.9\textwidth,clip=true,trim=0in
      8.5in 0.35in 0.035in]{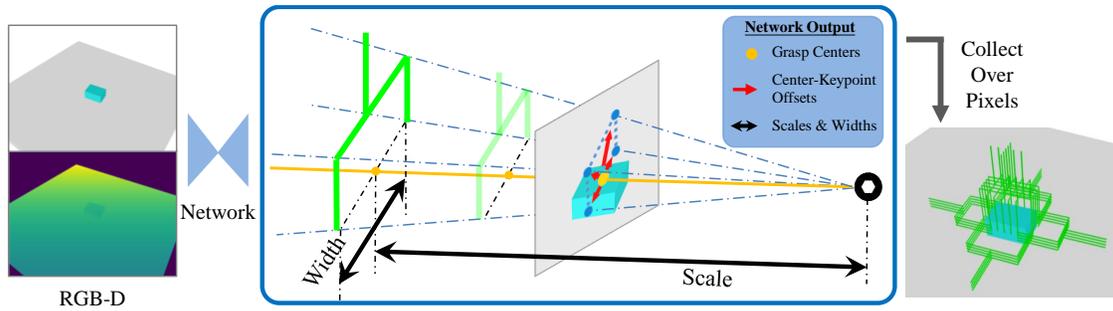}}};
    \end{tikzpicture}
  }
  \vspace*{-0.15in}
  \caption{
  \textbf{Overview of \netName.} 
  Given an RGB-D input, our model predicts pixel-wise candidate grasp poses.
  It estimates the \textit{pose up to a scale} by applying \PnP algorithm on generated image-space keypoint coordinates with camera intrinsic matrix.
  The keypoints are obtained by predicted grasp centers and offsets. 
  Then grasp \textit{scale} (the magnitude of translation) as well as the open width are inferred, which complete the grasp pose parameters.
  The final synthesized grasp set is the collection of results over high-confident pixels.
  }
  \vspace*{-0.2in}
 \label{fig:method}
\end{figure*}

\section{Methodology}
\label{sec:meth}

\begin{figure}[!ht]
  \vspace*{-0.1in}
    \centering
    \begin{minipage}{.5\linewidth}
        \centering
        \includegraphics[width=\textwidth, clip=true,trim=0in
      0in 0.5in 0in]{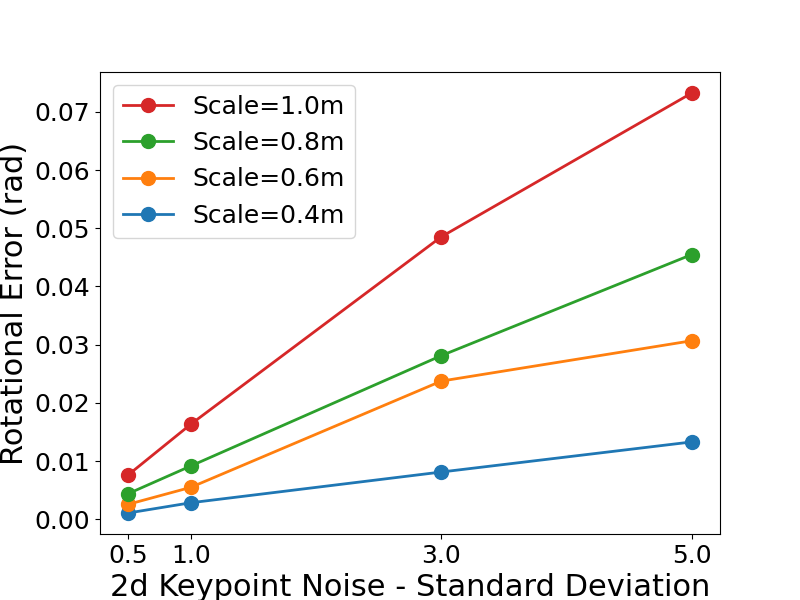}
        \label{fig:kptRot}
    \end{minipage}\hfill
    \begin{minipage}{0.5\linewidth}
        \centering
        \includegraphics[width=\textwidth, clip=true,trim=0in
      0in 0.5in 0in]{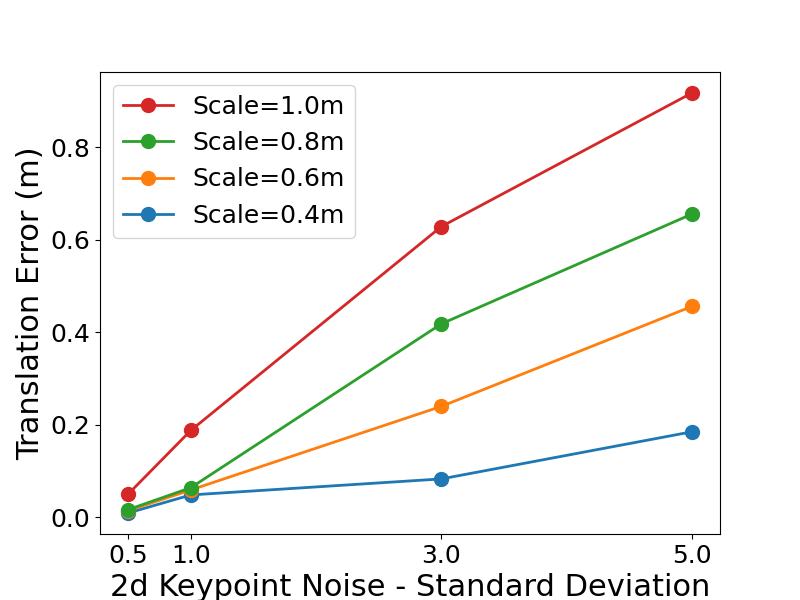}
        \label{fig:kptTrl}
    \end{minipage}
      \vspace*{-0.08in}
    \caption{Synthetic study on the relationship between pose scale and pose recovery error from \PnP algorithm \cite{ippe} due to noise.
     With larger grasp pose scale (grasp is further from the camera), both rotational and translational error increases under all noise levels. 
     The observation motivates us to predict scaled keypoint location as in \S \ref{sec:methPose}.
    }
    \label{fig:kptScale}
     \vspace{-0.15in}
\end{figure}

\subsection{Problem Definition}
\label{sec:methProb}

Given a monocular RGB-D input, our goal is to synthesize a set of 6-DoF grasps with grasp pose $\grasp \in \SE3$ and associated open width $\graspWidth$ that can pick up objects perceived by the image \textit{without} converting the input into 3D representations such as point cloud.
The problem is challenging since the input is in 2D image space while output is in 3D space.

Our method, as illustrated in Fig. \ref{fig:method}, separately predicts poses from keypoints, the scale of pose, and the open width.

\subsection{Pose Estimation with Scale-Normalized Keypoint}
\label{sec:methPose}

Inspired by KGN \cite{KGN}, we adopt a keypoint-based strategy that leverages well-established camera 3D-to-2D projection principle to estimate grasp poses up to a scale.
Specifically, given RGB-D input, \netName predicts a set of \textit{grasp centers} $\{\gCenter^\ori_i\}_i$, defined as the center point between gripper tips,
for each orientation interval of line segment between gripper tips in the image space: $\ori \in \{1, 2, \cdots, \oriNum \}$. 
The design of orientation interval is to enable simultaneous detection of multiple grasps that share the same center, which is useful for generating diverse candidate set for rotationally symmetric objects(e.g. grasps for a ball). It also functions as a non-maximum suppression mechanism by considering grasps with overlapping centers and similar orientations to be highly simillar, resulting in only one grasp being retained. 
The grasp centers are detected from heatmap $\centerHeatmap \in [0, 1]^{\width^\prime \times \height^\prime \times \oriNum}$, where $\width^\prime$ and $\height^\prime$ represents resolution of downscaled feature map.
As we will discuss soon, center-based strategy provides a simple way to group keypoints. It is also straightforward to fuse with features extracted from other modalities, such as language \cite{chen2021joint}.

With grasp centers, the keypoints' locations $\{(\pt^\ori_{i1},\pt^\ori_{i2}, \pt^\ori_{i3}, \pt^\ori_{i4} )\}$ are predicted based on offset estimation.
Our network learns to generate center-keypoint offset vector map $\centerKptsOffset$, which encodes the displacement from center to keypoints for each center and orientation.
The keypoints locations can be obtained from the centers and offsets, and are then fed into IPPE \cite{ippe} algorithm specifically designed for coplaner \PnP problem to produce 6-DoF grasp pose.
The final synthesized grasp set is the collection of results from grasp centers with high confidence.

\begin{figure}[t!]
 
  \vspace*{-0.0in}
  \centering
  \scalebox{0.97}{
    \begin{tikzpicture}
     \node[anchor=north west](figs) at (0in,0in)
      {{\includegraphics[width=0.9\linewidth,clip=true,trim=0in
      9.3in 3.8in 0.0in]{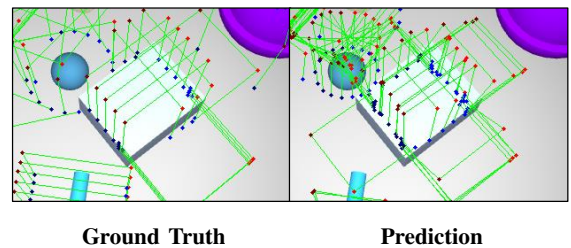}}};
     \node[anchor=north] at ($ (figs.south) + (-0.75in, 0.0) $) {\bf \small Ground Truth};
     \node[anchor=north] at ($ (figs.south) + (0.75in, 0.0) $) {\bf \small Prediction};
    \end{tikzpicture}
  }
  \caption{
  Example of inaccurate keypoints proximity prediction on primitive shape data.
  The predicted keypoints exhibit greater proximity to each other than the ground truth keypoints, possibly due to the influence of visual disturbance from surrounding objects, resulting in imprecise scale estimation.
  }
 \label{fig:inaccKpts}
 \vspace{-0.15in}
\end{figure}

\textbf{Scale-Normalized Keypoint Prediction}
A natural choice of keypoint design is to define pair-wise distance equal to open width,
which places two keypoints on gripper tips and other two above to form a square.
In this way, 2D keypoints on the image are adaptive to the size of graspable part.

However, such a design \textbf{naturally bias towards grasp pose with small \textit{scale} (in close proximity to camera) in training.}.
Unlike human pose or object pose estimation problem \cite{CenterNet2019},
grasp poses are larger in quantity per image due to multiple grasps feasible for each object, exhibiting a distribution of scales.
For fixed size grasps, the keypoints appear closer on the image for further poses as a result of perspective projection.
However, one property for \PnP algorithm is that \textit{recovered grasp pose for close-by points are prone to larger error under the same level of noise, as keypoint structure is more strongly disrupted.}
Hence, same error in Euclidean keypoint space transfers to larger error in pose space for large-scale grasp annotations.

To empirically confirm this property, we conduct a synthetic experiment to examine the relationship between scale and pose recovery error.
We randomly sample grasp orientation at the origin, and place cameras with various distance along optical axis. 
Then we recover gripper pose with canonical keypoint projections injected by gaussian noise, and calculate the average rotational and translational error, defined in the same way as in \S \ref{sec:expDataset}.
The result is shown in Fig. \ref{fig:kptScale}, which indicates that both errors increase as a function of scale conditioned on any noise level.

We propose to predict image-space keypoints normalized by the scale. 
The idea is related to human/object keypoints prediction with area-normalization \cite{DEKR, yoloPose} or object-size-normalization \cite{centerPose2022},
but for grasp pose estimation problem we normalize with pose proximity to camera center to introduce scale invariance.
Specifically, we scale the offset which determines the proximity of keypoints. For each grasp center $\gCenter$ and associated \textit{actual} offset vectors $\tilde{\centerKptsOffset}_\gCenter$ and scale $\scaleMap_\gCenter$, our network is tasked to predict: 
\begin{align*}
	\centerKptsOffset_\gCenter = \tilde{\centerKptsOffset}_\gCenter / \scaleMap_\gCenter
\end{align*}
where the predicted scale (see Sec. \ref{sec:methScale}) is used in the inference.
The scale-normalized keypoint design reduces the susceptibility of pose recovery to noise for further grasp poses.
Assuming the prediction of scaled offsets suffers from zero-mean Gaussian noise $\epsilon \sim \gauss(0, \sigma^2)$.
Then the perturbance of original offsets is reduce to $ (1 / \scaleMap_\gCenter) \epsilon \sim \gauss(0, \sigma^2 / \scaleMap_\gCenter^2)$, whose standard deviation is decreased by the scales, leading to more accurate pose recovery.
We show in Sec. \ref{sec:expDataset} that this design improves the pose estimation accuracy.

\begin{table*}[!ht]
  \vspace*{0.03in}
  \centering
  \setlength\tabcolsep{2.5 pt} 
  \renewcommand{\arraystretch}{1.3}
  \caption{Vision Dataset Evaluation}
  \begin{threeparttable}
  \begin{tabular}{|c| c  c  c  c  c  c  c  c  c | c  c  c  c c c c c c | }
  		\hline
	\multirow{2}{*}{Methods} &
    \multicolumn{9}{c|}{\textbf{Single-Object Evaluation (GSR\% / GCR\% / OSR\%) }} &
    \multicolumn{9}{c|}{\textbf{Multi-Object Evaluation (GSR\% / GCR\% / OSR\%) }} 
    \\ 
		\cline{2-19}
 	{} & \multicolumn{3}{c|}{$1\text{cm}+20^\circ$} &
    \multicolumn{3}{c|}{$2\text{cm}+30^\circ$} & 
    \multicolumn{3}{c|}{$3\text{cm}+45^\circ$} &
    \multicolumn{3}{c|}{$1\text{cm}+20^\circ$} &
    \multicolumn{3}{c|}{$2\text{cm}+30^\circ$} & 
    \multicolumn{3}{c|}{$3\text{cm}+45^\circ$} 
    \\ \hline
 	
%
%
%
%
 	Contact-Graspnet$\dagger$ & 
 	\multicolumn{3}{c|}{29.9 / 24.9 / 77.0 } &
 	\multicolumn{3}{c|}{60.1 / 32.0 / 81.7 } &
 	\multicolumn{3}{c|}{81.6 / 36.5 / 84.2} & 
 	\multicolumn{3}{c|}{22.1 / 15.5 / 44.1} &
 	\multicolumn{3}{c|}{54.2 / 28.5 / 51.4} &
 	\multicolumn{3}{c|}{78.4 / 34.5 / 54.4}   
 	\\
%
%
%
	\hline
	\hline 
	
	KGN	\cite{KGN} (single)\tnote{1} & 
 	\multicolumn{3}{c|}{55.5 / 42.9 / 97.0 } &
 	\multicolumn{3}{c|}{78.5 / 63.3 / 99.6 } &
 	\multicolumn{3}{c|}{86.9 / 73.2 / 99.9} & 
 	\multicolumn{3}{c|}{10.8 / 5.48 / 28.7} &
 	\multicolumn{3}{c|}{30.6 / 18.7 / 51.8} &
 	\multicolumn{3}{c|}{49.6 / 33.8 / 62.4}  
 	\\

	KGN \cite{KGN} (multi)\tnote{1}	&
 	\multicolumn{3}{c|}{38.6 / 18.5 / 63.7 } &
 	\multicolumn{3}{c|}{63.8 / 33.1 / 85.0 } &
 	\multicolumn{3}{c|}{78.4 / 46.2 / 91.0} & 
 	\multicolumn{3}{c|}{52.6 / 40.7 / 86.5} &
 	\multicolumn{3}{c|}{78.1 / 66.7 / 93.1} &
 	\multicolumn{3}{c|}{88.2 / 78.2 / 94.8}  
	\\

 	\hline
	\hline 
	
	\netName (single)\tnote{1}	&
 	\multicolumn{3}{c|}{81.4 / 59.1 / 98.8 } &
 	\multicolumn{3}{c|}{92.7 / 70.9 / 99.7 } &
 	\multicolumn{3}{c|}{96.0 / 77.4 / 99.8} & 
 	\multicolumn{3}{c|}{21.4 / 15.3 / 42.9} &
 	\multicolumn{3}{c|}{41.1 / 32.2 / 58.4} &
 	\multicolumn{3}{c|}{56.7 / 45.9 / 68.7}  
	\\
 		
 	\netName (multi)\tnote{1}	& 
 	\multicolumn{3}{c|}{\textbf{86.4} / \textbf{61.8} / \textbf{99.7} } &
 	\multicolumn{3}{c|}{\textbf{93.4} / \textbf{72.5} / \textbf{1.00} } &
 	\multicolumn{3}{c|}{\textbf{95.7} / \textbf{80.4} / \textbf{1.00} } & 
 	\multicolumn{3}{c|}{\textbf{80.4} / \textbf{58.5} / \textbf{93.1}} &
 	\multicolumn{3}{c|}{\textbf{91.0} / \textbf{73.5} / \textbf{94.6}} &
 	\multicolumn{3}{c|}{\textbf{95.1} / \textbf{80.5} / \textbf{94.9}}  
 	\\
 		\hline
  \end{tabular}
  \begin{tablenotes}
	  \item[1] Single and multi in the paratheneses means trained on single-object or multi-object data.
	  \item[$\dagger$] The evaluated model is trained on Acronym \cite{acronym2021} dataset.
  \end{tablenotes}
  \end{threeparttable}
  \label{tab:visResults}
  \vspace*{-0.15in}
\end{table*}

\begin{table}[ht!]
  \vspace*{0.00in}
  \centering
  \caption{Ablation Study Results}
  \renewcommand{\arraystretch}{1.3}

  \begin{threeparttable}
  \begin{tabular}{|c| c c | c  c | }
      \hline
	\multirow{2}{*}{Methods} & 
	\multicolumn{2}{c|}{Components} & 
	\multicolumn{2}{c|}{Mult-Object Evaluation} \\
  		\cline{2-5}
	{} & 
	sBranch\tnote{1} & sKpt\tnote{2} & 
	\multicolumn{2}{c|}{${}^\star$Avg GSR\% / GCR\% / OSR\%} 
	\\
 		\hline
 	
 	KGN & 
 	{} & {} &
 	\multicolumn{2}{c|}{30.4 / 19.3 / 47.6} \\
 	
 	\netName & 
 	\bluecheck & {} &
 	\multicolumn{2}{c|}{ 38.8 / 30.7 / 53.2} \\
 	
 	\netName & 
 	\bluecheck & \bluecheck &
 	\multicolumn{2}{c|}{\textbf{39.7} / \textbf{31.1} / \textbf{56.7}} \\
 	
    \hline
  \end{tabular}
  \begin{tablenotes}
	  \item ${}^\star$ Numbers averaged over three error tolerance levels.
	  \item [1] sBranch - Scaled branch (\S.\ref{sec:methScale}).
	  \item [2] sKpt - Scale-normalized keypoints (\S \ref{sec:methPose}). 
  \end{tablenotes}
  \end{threeparttable}
  \label{tab:ablate}
  \vspace*{-0.15in}
\end{table}

\subsection{Scale Prediction}
\label{sec:methScale}

Although keypoints with open width prediction are sufficient to recover full \SE3 pose from \PnP algorithm, 
empirically the generated scales are inaccurate as the proximity between keypoints cannot be stably predicted, especially under domain shifts.
For example, when testing on multi-object scenes from the Primitive Shape dataset \cite{KGN}, keypoint detector trained on single-object data tends to produce keypoints that are more closely grouped than the ground truth. 
Since the distance to the camera optical center is inversely proportional to the size of objects in the image, imprecise image-space proximity can lead to erroneous scale estimation.

KGN \cite{KGN} mitigates the problem by heuristically replacing the predicted scale with the perceived depth at the grasp center from the sensor.
That leads to two potential problems. 
First, center depth is not identical to grasp translation scale, as the center point is not observable surface point. When grasping a box, the gripper enclosing center would fall inside of the box, which makes it unperceivable. 
Hence, the heuristics will introduce additional error.
Furthermore, depth sensors may not always provide reliable measurements. 
Perceived depth maps can be affected by various sources of noise and may contain missing values \cite{depthFail}. As a result, the raw depth values may be less informative for accurately estimating the scale of a grasp pose.

Rather than conditioning the scale estimation on keypoints or raw depth value, \netName directly predicts a scale map $\scaleMap \in \Real^{\height^{\prime} \times \width^{\prime} \times \oriNum}$ for each pixel and orientation. 
Each grasp pose can be easily assocatiated with the scale prediction at the corresponding grasp center location and orientation interval.
Although similar to depth prediction problem, scale estimation given solely RGB input is essentially ill-posed, we assume the noisy depth map input serves as additional signal that can reduce the ambiguity.
With accurate scale predictions, the translation magnitude from the \PnP can be easily refined: Suppose for a predicted grasp center $\gCenter$ and an orientation interval $\ori$, the rotation and translation given by \PnP algorithm is:
$\tilde{g} = \{ \tilde{\rotMat}, \tilde{\trl} \}$, 
then the final pose combined with the scale prediction:

\begin{align*}
	\hat{g} & = \{ \hat{\rotMat}, \hat{\trl} \})  = \{ \tilde{\rotMat}, \frac{\scaleMap_{\gCenter, \ori}}{\norm{ \tilde{\trl}}} \tilde{\trl} \})
\end{align*}

\subsection{Final Loss}
\label{sec:methLoss}

The training of the network requires labels for all branch outputs, which can be generated easily given only annotation of grasp poses and camera intrinsic and extrinsic matrices.
The objective for training given ground truths involves 
the penalty-reduced focal loss for heatmap $\loss_\centerHeatmap$, 
plus the $L_1$ regression loss for center-keypoint offsets $\loss_\centerKptsOffset$, open width $\loss_\graspWidthMap$, and translation scale $\loss_\scaleMap$ on labeled grasp centers.
The final loss is the weight sum of the four losses:
\begin{align*}
	\loss = \gamma_\centerHeatmap \loss_\centerHeatmap  + \gamma_\centerKptsOffset \loss_\centerKptsOffset + \gamma_\graspWidthMap \loss_\graspWidthMap + \gamma_\scaleMap \loss_\scaleMap
\end{align*}
In this paper we choose: $\gamma_\centerHeatmap=1$, $\gamma_\centerKptsOffset=1$, $\gamma_\graspWidthMap=10$, $\gamma_\scaleMap=10$.


\section{Experiments}
\label{sec:exp}

\subsection{Synthetic Dataset}
Following \cite{KGN}, we use the Primitive Shape (PS) dataset for training our network.
The dataset is a synthetic dataset generated by spawning objects of simple shapes with random pose on the tabletop, which is the most common evaluation scenerio for grasp detectors.
The shapes we use involves: \textit{Cylinder}, \textit{Ring}, \textit{Stick}, \textit{Sphere}, \textit{Semi-sphere}, and \textit{Cuboid}.
Ground truth grasps are annotated by sampling evenly distributed instances from \textit{grasp families} \cite{ps2022} - the closed-form grasp pose distributions parameterized by the shape type and sizes, which is assumed to sufficiently cover the feasible grasp modes for a given primitive shape based on human expertise due to the simplicity of the object geometry.

We choose the dataset since shape decomposition proves to be a very effective strategy that drives force in grasp synthesis research for years \cite{ps1, ps2, ps2022, ps3}.
The grasp label generation approach is significanly more cost-effective compared to sample-then-verify strategy based on simulators \cite{acronym2021}, \cite{jacquard2018}, \cite{GPNet2020}.
Furthermore, it mitigates potential bias or inaccuracies in the labeling process that can arise from sampling artifacts \cite{sampleInvest}, thereby avoiding negative impact on training or evaluation process.

The PS dataset contains 1000 \textit{single-object} scenes, divided into 800 training scenes and 200 test scenes. 
For each scene, RGB-D data is rendered from 5 random camera poses, leading to 4000 training data and 1000 test data.
\textit{In addition}, we generate a multi-object PS dataset of the same quantity, where all 6 shapes with random size and color are spawned in each scene and the grasp poses causing collisions are removed.
We increase the annotation sample density for test splits to verify the extrapolation ability of trained grasp detector from sparse examples.

\subsection{Implementation Details}
The vision encoder used in our method is DLA-34 \cite{dla} modified with deformable convolution layer \cite{dcn}, except that the first layer is modified with 4-channel kernels for RGB-D input.
A shallow two-layer convolution network is used for each task head.
We finetune the network on PS training splits starting from pretrained weights on CoCo dataset \cite{coco}, whose blue channel parameters are duplicated for the depth channel in the input layer.
The network is trained for 400 epochs using the ADAM optimizer, with initial learning rate as $1.25 \times 10^{-4}$ and is decayed by $10$x at epoch 350 and 370.
We adopts image augmentation, including random cropping, flipping, and color jittering, for better generalization ability.
All the training on done on a single NVIDIA RTX 3090 GPU, and testing on a single NVIDIA RTX 1080Ti.
The training takes 16 hours, and the inference speed is 9 FPS.

\subsection{Synthetic Dataset Experiments}
\label{sec:expDataset}

We first test our method on the Primitive Shape dataset test split to exmaine its ability to learn the annotated grasp distribution.
The performance is compared against KGN \cite{KGN} to demonstrate efficacy of proposed modifications.
We further conduct ablation study to break down the contribution by each proposed components under domain shift.

\textbf{Metrics: }
We evaluate predicted grasp pose set by comparing against the ground truth (GT) set. 
Following \cite{KGN}, we use three metrics for the evaluation: 
(1) \textit{Grasp Precision Rate (GPR)}: Percentage of grasp predictions with closeby GT;  
(2) \textit{Grasp Coverage Rate (GCR)}: Percentage of GT pose with closeby predictions;
(3) \textit{Object Success Rate (OSR)}: Percentage of objects targeted by near-GT predictions.
The similarity between two poses are determined by thresholding both the translational and rotational errors, 
which are defined as $L_2$ norm between translations and the minimum angle required to align rotations. \cite{hartley2013rotation}, \cite{localBenchmark}, respectively.
We collect the results under three different error levels from strict to loose:
$(1 \text{cm}, 20^{\circ})$, $(2 \text{cm}, 30^{\circ})$,
and $(3 \text{cm}, 45^{\circ})$.

\textbf{Dataset Evaluation Results:}
We first evaluate \netName and baseline KGN on both single- and multi-object test sets, while training both methods on either single- or multi-object training sets.
The results are tabulated in Tab. \ref{tab:visResults}, which includes the performance of Contact-GraspNet trained on clutter scenes from Acronym \cite{acronym2021} for reference.
We first notice that \textit{\netName outperforms the baseline significantly under all settings.}
For example, when trained and tested both on multi-object data, \netName achieves $27.8\%$, $17.8\%$, and $6.6\%$ performance improvement under the most sctrict threshold values for GSR, GCR, and OSR, respectively.
When trained on snigle-object scenes and generalizing to more complex multi-object scenarios, \netName gets $10.6\%$, $9.8\%$, and $14.2\%$ performance gain for the above metrics, which is comparable to Contact-Graspnet considered as an upper bound under this setting \cite{KGN}.

We also observe that our method trained on multi-object data performs even better in single-object benchmark compared to it trained also on single-object scenarios. 
This suggests that our network learns to reason about physical relationship between the objects, which is beneficial for all grasping tasks including single-object picking. 
A similar trend is not observed for \cite{KGN} - trained on multi-object data, its precision in single-object evaluation is 
$47.8\%$ lower than that of \netName under the most strict error tolerance thresholds in single-object testing, and is also $14\%$ lower than itself in multi-object object testing.
The fact that it cannot generalize to even simpler task suggests deficit in its design that prevents it from reasoning with geometric information, which is mitigated by our modifications.

\begin{figure}[!t]
 
  \vspace*{-0.0in}
  \centering
  \scalebox{0.75}{
    \begin{tikzpicture}
     \node[anchor=north west] at (0in,0in)
      {{\includegraphics[width=0.8\linewidth,clip=true,trim=0in
      7.8in 4.6in 0.035in]{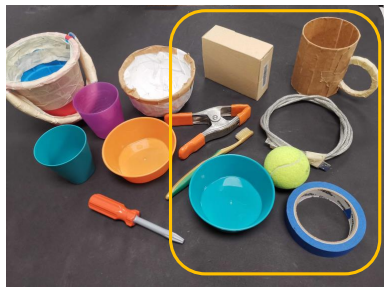}}};
    \end{tikzpicture}
  }
  \vspace*{-0.25in}
  \caption{Objects used for physical experiments.
  Yellow bounding box selects the object set for single-object grasping.
  }
 \label{fig:phyObjs}
 \vspace{-0.2in}
\end{figure}

\begin{figure*}[th!]
  \vspace*{0.0in}
  \centering
  \scalebox{0.98}{
    \begin{tikzpicture}
     \node[anchor=north west](figs) at (0in,0in)
      {{\includegraphics[width=\linewidth,clip=true,trim=0in
      9.6in 0.0in 0.035in]{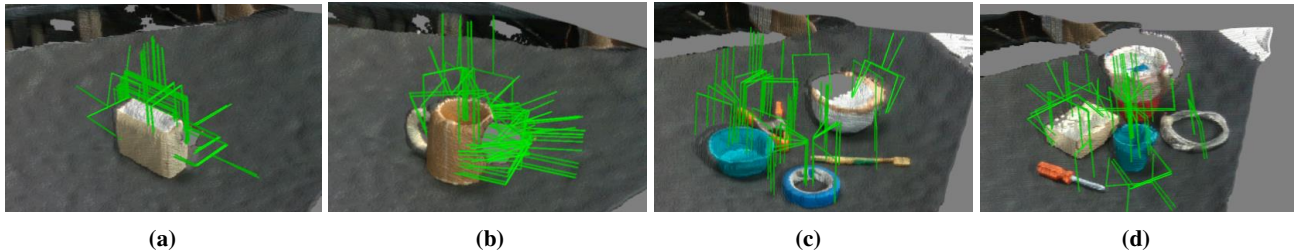}}};
     \node[anchor=north] at ($ (figs.south) + (-2.6in, 0.0in) $) {\bf
     \small (a)};
     \node[anchor=north] at ($ (figs.south) + (-0.85in, 0.00in) $) {\bf
     \small (b)};
     \node[anchor=north] at ($ (figs.south) + (0.85in, 0.00in) $) {\bf \small
     (c)};
     \node[anchor=north] at ($ (figs.south) + (2.6in, 0.0in) $) {\bf
     \small (d)};
    \end{tikzpicture}
  }
  \vspace*{-0.2in}
  \caption{Demonstration of generated grasp candidates in physical experiments.
  (a)(b) Single-object experiment results.
  (c)(d) Multi-object experiment results.
Only at most 40 grasps are randomly selected for visualization.
  }
 \label{fig:phyDemo}
 \vspace*{-0.2in}
\end{figure*}

\textbf{Ablation study:}
To better understand the benefits of our introduced upgrades, we conduct ablation study by removing the scale-normalized keypoint and scale prediction branch design one-by-one.
To examine the capability of grasp detector under domain shift, the networks are trained on single-object data while tested in multi-object environments.
We report the average number of three metrics under all error tolerance levels in Tab. \ref{tab:ablate}.
The result demonstrates the contribution by both modifications -
simple scale branch dramatically improves the performance, while scale-normalized keypoint further enhances the accuracy.

\subsection{Physical Experiments}
To validate the sim-to-real generalization ability, we apply the proposed grasp detector in real-world physical experiments.
The robotic system is composed of an Intel RealSense D435 camera mounted at a fixed position for perception, and a custom-made 7-DoF robotic arm for execution.
The trajectory is planned with MoveIt \cite{moveit}.
The object set used for experiments is depicted in Fig. \ref{fig:phyObjs}.
Both \textit{single-object} and \textit{multi-object} grasping experiments are performed.
For all physical experiments, we use the \netName weight trained on Primitive Shape multi-object training set, as it demonstrates superior performance even in single-object vision dataset evaluation.
$95\%$ confidence interval are reported.

\begin{table}[t]
        \centering
          \caption{Single-Object Grasping Comparison from Published Works
    \label{tab:phySingle}}
      \renewcommand{\arraystretch}{1.2}
        \begin{threeparttable}
          \setlength\tabcolsep{7pt}
          \begin{tabular}{|l|c|c|c|c|}
          \hline
          \multicolumn{1}{|l|}{\multirow{2}{*}{\bf{Approach}$^\star$ }}  &
            \multicolumn{3}{c|}{\multirow{2}{*}{\bf{Settings}}} &
            \multicolumn{1}{c|}{\bf{Success}} \\
            \multicolumn{1}{|c|}{}                   &
            \multicolumn{3}{c|}{}                   &
            \multicolumn{1}{c|}{\bf{Rate (\%)}} \\ 
            \hline
           & \multicolumn{1}{c|}{Modality} 
           & \multicolumn{1}{c|}{Obj}
           & \multicolumn{1}{c|}{Trial/Obj} 
           & 
             \\ \hline
         PointNetGPD\cite{pointnetGPD2019}   & PC  & 10  &  10  & 82.0
               \\ \hline
          6DoF-GraspNet\cite{6dofGraspNet2019} & PC & 17  &  3 & 88.0

                     \\ \hline
        L2G\cite{L2G2022}  & PC &     48  & 5  & 50.5

        	\\ \hline \hline
          RGBMatters\cite{rgbGrasp2021} & RGB-D  &  9 & 20 & 91.67
           \\ \hline  
          MonoGraspNet\cite{monoGrasp} & RGB  &  12  & 15 & 75.95
            \\ \hline \hline 
          KGN  &  RGB-D  & 8  & 5 & 87.5$\pm$9.6 
          \\ \hline
          KGNv2  &  RGB-D  & 8  & 5 & 92.5$\pm$6.7
            \\ \hline 
        \end{tabular} 
        \begin{tablenotes} 
          \footnotesize 
          \item ${}^\star$ All results for baselines adopted from original paper for reference, following \cite{gknet2021}.
        \end{tablenotes}
        \end{threeparttable}
        \vspace*{-0.15in}
\end{table}

\textbf{Grasp selection strategy.}
A grasp selection strategy is necessary to choose a pose for execution from the rich candidate set generated by \netName.
Furthermore, the grasp poses in dataset are annotated in a gripper-agnostic way without considering specific gripper geometry.
Hence, gripper-specific prior is injected in the selection stage.
To select the pose for execution, we first rank the candidate poses based on method fidelity.
We calculate a score for each grasp pose $s(\grasp)$ by combining the center confidence generated during the keypoint detection stage with the reprojection error (RE) introduced by the pose recovery stage: $s(\grasp) = \centerHeatmap_{\gCenter, \ori} + RE$.
Then, we choose the feasible pose with top confidence that causes no collision and encloses non-empty volume of point cloud based on gripper attributes \cite{GPD2017}.
For experiment outcome, we collect numbers from related papers as reference.

\begin{table}[t]
        \centering
          \caption{Multi-Object Grasping Comparison from Published Works
  	\label{tab:phyMulti}}
      \renewcommand{\arraystretch}{1.0}
        \begin{threeparttable}
          \setlength\tabcolsep{4.5pt}
          \begin{tabular}{|l|c|c|c|c|c|}
          \hline
          \multicolumn{1}{|l|}{\multirow{2}{*}{\bf{Approach}${}^\star$}}  &
            \multicolumn{3}{c|}{\multirow{2}{*}{\bf{Settings}}} &
            \multicolumn{1}{c|}{\bf{Success}} 
            & \multicolumn{1}{c|}{\bf{Clear}}  \\
            \multicolumn{1}{|c|}{}                   &
            \multicolumn{3}{c|}{}                   &
            \multicolumn{1}{c|}{\bf{Rate (\%)}} &
            \multicolumn{1}{c|}{\bf{Rate (\%)}} \\ 
            \hline
           & \multicolumn{1}{c|}{Modality} 
           & \multicolumn{1}{c|}{Sn${}^\dagger$}
           & \multicolumn{1}{c|}{Obj/Sn${}^\dagger$} 
           &  & 
             \\ \hline
         PnGPD\cite{pointnetGPD2019}   & PC  &  10  &  8 & 77.77  & 97.5
                     \\ \hline
        CGN\cite{ContactGraspNet2021} & PC  &  9  &  4-9  & 90.20  & N/A$\ddagger$ 
        \\  \hline
         Pn++\cite{Pointnet++2017} & PC  &  20  &  6  & 77.19  & 94.5

        	\\ \hline \hline
          RGBMatters\cite{rgbGrasp2021} & RGB-D  & 6  & 5-8 & 91.1 & 100
           \\ \hline  
          MonoGN\cite{monoGrasp} & RGB  & 8 & 4-5 & N/A$\ddagger$ & 80.6
            \\ \hline \hline 
          KGNv2  &  RGB-D  & 10 & 5 & 80$\pm$10.1 & 96$\pm$7.4
            \\ \hline 
        \end{tabular} 
  \begin{tablenotes}
  	  \footnotesize 
	  \item ${}^\star$ All results for baselines adopted from original paper for reference, following \cite{gknet2021}.
	  \item $\dagger$ Calculated as the average success number over average attempts number.
	  \item $\ddagger$ Not released by original paper
  \end{tablenotes}
        \end{threeparttable}
        \vspace*{-0.15in}
\end{table}

\textbf{Single-Object Grasping Results.}
We conduct experiments on pick-and-place task of individual objects, requiring the robotic arm to successfully retrieve a target that has been randomly placed, and transport it to a predetermined location.
In this experiment, we utilize the same set of eight objects with diverse shapes as those employed in \cite{KGN}, shown in Fig. \ref{fig:phyObjs}.
For each object, we conduct 5 trials and calculate the success rate.

The results are collected in Table. \ref{tab:phySingle}.
Following \cite{gknet2021}, \cite{ps2022}, we also collect the results from published grasping research efforts to place the performance of \netName within greater context.
\netName demonstrates a satisfacory success rate in spite of trained on basic, synthetic primitive shapes, indicating that geometric information is learnt.
Furthermore, it achieves a $5\%$ performance gain compared to KGN, suggesting the proposed modifications lead to more accurate grasp pose prediction.
Typical cause of failure is the prediction of unstable grasps for target objects, involving off-center grasp pose for the ball causing it to roll off, or targeting metallic, slippery section on the clamp.

\textbf{Multi-Object Grasping Results}
In addition, we conduct experiments in scenarios with multiple objects, where five objects are randomly selected and placed on the table for each scene. 
We iteratively select grasp poses generated by the model for execution. 
The termination criteria for each trial consists of two conditions, namely:
(1) succesful removal of all objects;
(2) three consecutive failed attempts,
with the purpose of preventing the system from becoming stuck in a consistent failure mode.

We evaluate the succcess rate for grasp attempts and clearance rate for the objects.
The experiment results are tabulated in Tab. \ref{tab:phyMulti}, which collects results from related papers as before for reference. 
Our approach obtains a comparable success rate and clearance rate as state-of-the-art methods, which further validates our method in real-world tasks.
For failure cases, the cause for single-object picking failure still exist. 
In addition, we observe some failure where the grasps aim for occluded area, probably due to unreasonable extrapolation by the network.

\section{Conclusion}
\label{sec:conclusion}

In this work, we present a 6-DoF grasp pose detection method from RGB-D image input.
Our method first generates pose up to a scale based on image-space keypoint detection and \PnP algorithm. 
Then it regresses pose scale as well as open width.
Based on numerical analysis on PnP algorithm, we further propose the scale-normalized keypoints design to improve the pose estimation accuracy.
On the Primitive Shape dataset, we verify that our method learns to generate grasp distribution from labels better, and demonstrate the efficacy of designed components via ablation study.
Physical experiments are conducted to further validate our approach's generalization ability over sim-to-real gap.

While the physical experiments show that \netName successfully learns geometric reasoning skills that is generalizable to a set of common household objects from simple primitive geometric data, 
the uniform color of the primitive shapes may limit the model's capacity to recognize diverse visual appearances in the open world.
Future efforts could explore augmenting the dataset with authentic textures leveraging generative methods such as diffusion model \cite{zhang2023adding, richardson2023texture}.

\bibliographystyle{IEEEtran}
\bibliography{reference.bib}

\end{document}